\pdfoutput=1

\documentclass[11pt]{article}

\usepackage[preprint]{acl}

\usepackage{times}
\usepackage{latexsym}

\usepackage[T1]{fontenc}

\usepackage[utf8]{inputenc}

\usepackage{microtype}

\usepackage{inconsolata}

\usepackage{graphicx}

\usepackage{fontawesome5}

\usepackage{booktabs}
\usepackage{multirow}
\usepackage{float}
\usepackage[normalem]{ulem}

\usepackage[all]{nowidow}

\usepackage{hyphenat}

\usepackage{easyReview}

\title{Iterative Layer Pruning for Efficient Translation Inference}

\author{Yasmin Moslem* \vspace{3pt} \\
    ADAPT Centre \\
    Trinity College Dublin \\
    Dublin, Ireland \\
    {\footnotesize yasmin.moslem@adaptcentre.ie}
    \\\And
    Muhammad Hazim Al Farouq* \vspace{3pt} \\
    Kreasof AI \\
    Research Labs \\
    Jakarta, Indonesia \\
    {\footnotesize muhammad.hazim@kreasof.my.id}
    \\\And
    John D. Kelleher \vspace{3pt}\\
    ADAPT Centre \\
    Trinity College Dublin \\
    Dublin, Ireland \\
    {\footnotesize john.kelleher@adaptcentre.ie}
  }

\begin{document}
\maketitle

\def\thefootnote{{\scalebox{1.5}{*}}}
\footnotetext{These authors contributed equally to this work.}\def\thefootnote{\arabic{footnote}}

\begin{abstract}
\nohyphens{
Large language models (LLMs) have transformed many areas of natural language processing, including machine translation. However, efficient deployment of LLMs remains challenging due to their intensive computational requirements. In this paper, we address this challenge and present our submissions to the \textit{Model Compression} track at the Conference on Machine Translation (WMT 2025). In our experiments, we investigate iterative layer pruning guided by layer importance analysis. We evaluate this method using the Aya-Expanse-8B model for translation from Czech to German, and from English to Egyptian Arabic. Our approach achieves substantial reductions in model size and inference time, while maintaining the translation quality of the baseline models.
}
\end{abstract}

\section{Introduction}

Large language models (LLMs) have demonstrated powerful capabilities in diverse natural language processing tasks, including translation. However, LLMs are often computationally intensive, making them impractical to deploy in real-world settings with limited resources. To enhance the efficiency of these models, researchers have explored various model compression techniques, aiming to reduce their computational requirements while preserving quality \citep{Gandhi2023-Distil-Whisper,Sajjad2023-EffectLayerDropping,Treviso2023-Efficient-NLP,Sreenivas2024-LLMPruningMinitron,Gu2025-Jet-NemotronNeuralArchitectureSearch,Moslem2025-EfficientSpeech-IWSLT}.

Aya Expanse is an open-weight large language model with multilingual capabilities. The WMT 2025 Model Compression track \citep{Gaido2025-ModelCompressionFindings} required all submissions to be derived from the Aya-Expanse-8B model. 
This work focuses on translation from Czech to German and from English to Egyptian Arabic.

Our experiments build on established work on iterative layer pruning guided by layer importance evaluation \citep{Peer2022-GreedyLayerPruning,Moslem2025-EfficientSpeech-IWSLT}.
We apply iterative layer pruning to the baseline model Aya-Expanse-8B\footnote{\scriptsize\url{https://hf.co/CohereLabs/aya-expanse-8b}} which originally consists of 32 layers and 8.03B parameters. This approach incrementally identifies and removes layers with minimal contribution to translation quality, one layer at a time. To this end, we conduct layer importance evaluation by measuring translation performance without each layer. After identifying and removing the least critical layer, we repeat the layer importance evaluation on the remaining layers until reaching our pruning target. The pruned model resulting from this process is then fine-tuned on the News Commentary dataset. We have made three submissions; the primary submission is a 24-layer model with 6.28B parameters, and the two contrastive submissions are 20-layer and 16-layer models, with 5.41B and 4.54B parameters, respectively.

\section{Data}
\label{sec:data}

After layer pruning of the \textit{Aya-Expanse-8B} model (cf.~Section~\ref{sec:experiments}), we need to fine-tune the pruned model on medium-sized training data to restore the translation quality of the baseline model. 
To this end, we use the News Commentary dataset\footnote{\scriptsize\url{https://data.statmt.org/news-commentary/v18/training/}} which consists of news articles and their corresponding translations in several languages, including Arabic, English, German, and Czech.

We start by rule-based filtering of the Czech-to-German (CES-DEU) News Commentary dataset by removing duplicates, segments longer than 200 words, and those whose source/target length ratio is larger than 1.5 times. We also apply language detection with fastText\footnote{\scriptsize In particular, we used the fastText \textit{“lid.176.bin”} model.} \cite{Joulin2017-fastText} with a 0.9 threshold. Finally, we conduct semantic filtering using the \textit{mUSE} model \cite{Yang2020-mUSE} and \textit{Sentence-Transformers} \cite{Reimers2019-SentenceTransformers} with a 0.7 threshold of semantic similarity between the source and target. The CES-DEU News Commentary dataset includes 250.4K segments before filtering, and 201.3K segments after filtering.\footnote{\scriptsize\url{https://hf.co/datasets/ymoslem/news-commentary-cs-de}} Eventually, we split the dataset into train and test splits, where the test set includes 500 segments used for both testing and layer importance evaluation. Then, we sample 100K of the training data, using 0 as the random seed in both cases.

As the English-to-Arabic News Commentary dataset uses Standard Arabic,\footnote{\scriptsize\url{https://hf.co/datasets/ymoslem/news-commentary-en-ar}} we first apply the same rule-based and semantic filtering steps as those we employ while processing the Czech-to-German dataset, which result in 84.3K segments. Afterwards, we convert Standard Arabic text segments into Egyptian Arabic (ARZ) with GPT-4.1-Mini, using the prompt in Appendix~\ref{app:egy-prompt}, providing a fixed verified example that includes the English source as well as both the Standard Arabic and Egyptian Arabic translations. For parameters, we use temperature 0.3 and top-p 1. After completing the generation of the synthetic Egyptian Arabic translations, we apply rule-based filtering, comparing the generated Egyptian Arabic text segments to the original English source.
Finally, we calculate the semantic similarity between the English source and the Egyptian Arabic target and select the 500 segments with the highest scores (0.91-0.98) for the ``test" split, while the ``train" split comprises the remaining 83.2K segments.\footnote{\scriptsize\url{https://hf.co/datasets/ymoslem/news-commentary-eng-arz}}
Using this synthetic dataset to fine-tune our models yields clear quality gains compared to the baseline models, when evaluated on both the in-domain holdout test dataset (cf.~Table \ref{tab:eval-transformers}) and the WMT24++ benchmark\footnote{\scriptsize\url{https://hf.co/datasets/google/wmt24pp}} \cite{Deutsch2025-WMTt24} which includes 998 segments (cf.~Table \ref{tab:eval-arz-wmt}).

\begin{table*}[htbp]
\centering
\begin{tabular}{@{}llccc@{\hspace{10pt}}|cc@{}}
\toprule
\textbf{Language} & \textbf{Model} & \textbf{Layers} & \textbf{chrF++} ↑ & \textbf{COMET} ↑ & \textbf{Params (B)} ↓ & \textbf{Speed (mm:ss)} ↓ \\ \midrule
\multirow{4}{*}{\textbf{CES-DEU}} & Baseline & 32 & \textbf{52.79} & \textbf{87.18} & 8.03 & 00:47 \\ \cmidrule(l){2-7} 
 & \multirow{3}{*}{Pruned + FT}              & 24 & \uline{51.35} & \uline{85.70} & 6.28 & 00:34 \\
 &                                           & 20 & 49.45 & 83.95 & 5.41 & \textbf{00:27} \\
 &                                           & 16 & 45.79 & 79.39 & \textbf{4.54} & \textbf{00:27} \\ \midrule
\multirow{4}{*}{\textbf{ENG-ARZ}} & Baseline & 32 & 42.03 & 81.45 & 8.03 & 01:22 \\ \cmidrule(l){2-7} 
 & \multirow{3}{*}{Pruned + FT}              & 24 & \textbf{58.38} & \textbf{85.74} & 6.28 & 00:54 \\
 &                                           & 20 & 55.69 & 84.50 & 5.41 & 00:51  \\
 &                                           & 16 & \uline{51.17} & \uline{82.10} & \textbf{4.54} & \textbf{00:42}  \\ \bottomrule
\end{tabular}
\caption{Evaluation of layer pruning experiments. For translation from Czech to German (CES-DEU), pruning 8 layers and then fine-tuning the resulting model retains 98\% of the translation quality (as measured by COMET). Interestingly, for translation from English to Egyptian Arabic (ENG-ARZ), the model resulting from pruning up to 16 layers and then fine-tuning outperforms the Aya-Expanse-8B baseline for this language pair.}
\label{tab:eval-transformers}
\end{table*}


\section{Iterative Layer Pruning}
\label{sec:experiments}

As previous research demonstrates, iterative layer pruning achieves better quality than middle layer pruning \cite{Moslem2025-EfficientSpeech-IWSLT}. In this experimental setup, we apply iterative layer pruning to the \textit{Aya-Expanse-8B} baseline model. This approach incrementally identifies and removes layers with minimal contribution to translation quality, one layer at a time. The pruned models resulting from this process are then fine-tuned on the training dataset.
Furthermore, knowledge distillation data from the teacher model can be added.
Fine-tuning the pruned model restores most of the baseline model's translation quality. The following points elaborate on the process.

\paragraph{Layer importance evaluation:} We conduct layer importance evaluation by measuring translation performance without each layer. In this greedy layer pruning approach \citep{Peer2022-GreedyLayerPruning,Rostami2024-CULL-MT,Moslem2025-EfficientSpeech-IWSLT}, to prune \(n + 1\) layers, only a single optimal layer to prune must be added to the already known solution for pruning $n$ layers. After identifying and removing the least critical layer, we repeat the layer importance evaluation on the remaining layers until reaching our $n$ pruning target. We observe that while removing certain layers of the model (e.g. the first or last layer) substantially degrades translation performance, others result in minimal performance drops. Following \citet{Moslem2025-EfficientSpeech-IWSLT}, we use the chrF++ metric for layer importance evaluation for both better efficiency and quality.

\paragraph{Layer pruning:} We iteratively prune one decoder layer at a time, selecting the layer whose removal has the least negative impact on translation quality, measured by chrF++ scores. At each iteration, we evaluate the translation performance of the pruned model on the test split of the News Commentary dataset, after removing each candidate layer. The layer whose removal yields the best performance is eventually pruned. This process continues until a predefined number of layers (8, 12, and 16 layers) have been removed. By iteratively removing the least important layers, this performance-guided method produces a more compact model that can be fine-tuned further to recover the translation quality of the original model. We observe that the performance of the CES-DEU model is more impacted by pruning than the ENG-ARZ model, which might be attributed to the pre-training process (cf.~Table \ref{tab:eval-transformers}). In other words, the evaluation of the baseline for CES-DEU translation achieves better results than that for ENG-ARZ translation; hence, it seems that fine-tuning the pruned ENG-ARZ models has helped with improving the translation quality of this language pair.

\paragraph{Fine-tuning:} The pruning step is followed by fine-tuning the pruned model for 1 epoch using the News Commentary dataset (cf.~Section~\ref{sec:data}). The training uses a learning rate of 2e-5, a batch size of 8, and early stopping with a patience value of 5 evaluation runs, and it is conducted on one A100 80GB GPU.
This fine-tuning step recovers most of the translation quality of the baseline model.

\def\ci#1{\textcircled{\resizebox{.5em}{!}{#1}}}
\def\cis#1{\textcircled{\resizebox{.4em}{!}{#1}}}
\def\ciCheck{\scalebox{0.9}{\ci{\faCheck}}}
\def\ciTimes{\scalebox{0.9}{\cis{\faTimes}}}

\begin{table}[htbp]
\centering
\resizebox{\columnwidth}{!}{%
\begin{tabular}{@{}lcccc@{}}
\toprule
\textbf{Model} & \textbf{Layers} & \textbf{KD} & \textbf{chrF++} ↑ & \textbf{COMET} ↑ \\ 
\midrule
Baseline 32B & 40 & - & 54.57 &	87.76 \\
Baseline 8B  & 32 & - & 52.79 & 87.18 \\ 
\midrule
\multirow{6}{*}{Pruned + FT} & \multirow{2}{*}{24} & \ciTimes & 51.35 & 85.70 \\
                             &                     & \ciCheck & \uline{52.68} & \uline{86.50} \\ \cmidrule{2-5}
                             & \multirow{2}{*}{20} & \ciTimes & 49.45 & 83.95 \\
                             &                     & \ciCheck & \uline{51.25} & \uline{85.19} \\  \cmidrule{2-5}
                             & \multirow{2}{*}{16} & \ciTimes & 45.79 & 79.39 \\
                             &                     & \ciCheck & \uline{48.60} & \uline{81.39} \\
\bottomrule
\end{tabular}
}
\caption{Evaluation of knowledge distillation (KD). Fine-tuning pruned models on a combination of authentic and synthetic data (generated by Aya-Expanse-32B) improved the CES-DEU translation quality, with the \mbox{24-layer} pruned model nearly matching the performance of the Aya-Expanse-8B baseline.}
\label{tab:eval-kd}
\end{table}


\paragraph{Knowledge distillation:} To improve the quality of the CES-DEU models, we employed sequence-level knowledge distillation, where the student model is fine-tuned on a combination of authentic data and synthetic data generated by the teacher model for the same training dataset. In this case, the teacher model is the Aya-Expanse-32B while the students are the pruned models. After generating the data, we filter it by removing duplicates (exact matches in the target side of the authentic data), and translations with less than 70\% COMET scores, resulting in extra 98.6K segments of training data (cf.~Section~\ref{sec:data}). As Table~\ref{tab:eval-kd} demonstrates, fine-tuning the pruned models with a combination of both the authentic and knowledge distillation data has improved their translation quality, and helped close the performance gap between the 24-layer pruned model and the Aya-Expanse-8B baseline. Similarly, the 20-layer and 16-layer models show 2-3 points of improvement in terms of chrF++ and COMET metrics.

\section{Inference and Evaluation}

For inference, we use greedy generation by disabling the sampling options, and setting the temperature argument to 0. We apply a simple translation prompt: "Translate the following text from \{source\_language\} to \{target\_language\}:"

To evaluate our systems, we calculated BLEU \citep{Papineni2002-BLEU},  chrF++ \citep{Popovic2017-chrF++}, as implemented in the sacreBLEU library\footnote{\scriptsize\url{https://github.com/mjpost/sacrebleu}} \citep{Post2018-sacreBLEU}. For semantic evaluation, we use COMET \citep{Rei2020-COMET}.\footnote{\scriptsize In particular, we used the \textit{“wmt22-comet-da”} model.} Table~\ref{tab:eval-transformers} reports the results of the main experiments using the \textit{Transformers} framework\footnote{\scriptsize\url{https://github.com/huggingface/transformers}} \citep{Wolf2020-Transformers} for inference.

\begin{figure*}[ht]
    \centering
    \includegraphics[width=0.70\linewidth]{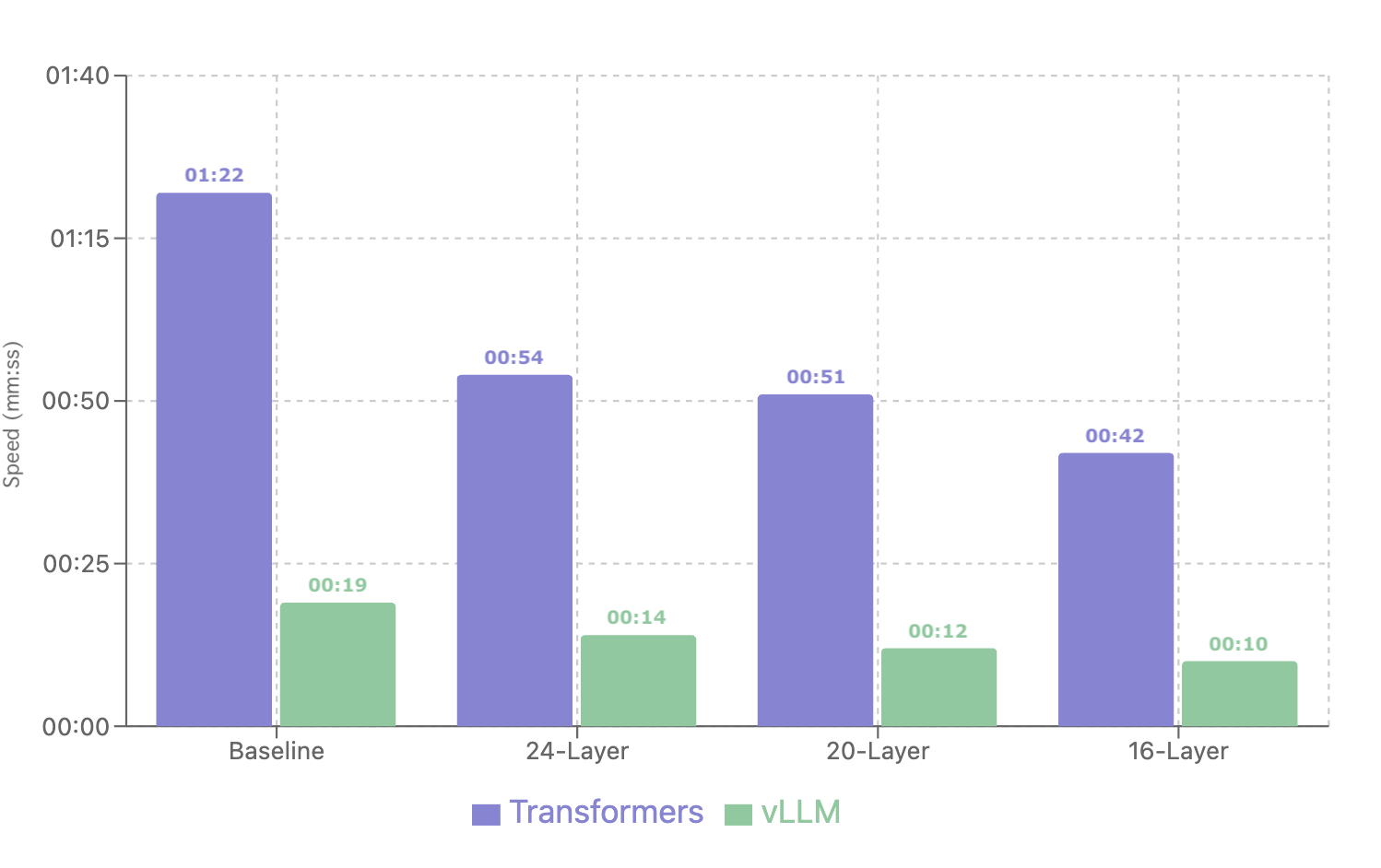}
    \caption{Inference speed comparison between \textit{Transformers} and \textit{vLLM}, using the Aya-Expanse-8B model for ENG-ARZ translation. \textit{vLLM} consistently outperforms \textit{Transformers} across all model sizes. Speedup ranges from 4.2x (16-layer) to 4.3x (baseline model). Both frameworks show improved performance with layer pruning. The 16-layer model achieves the fastest inference times overall.}
    \label{fig:trasformers-vllm}
\end{figure*}

\section{Results}

The process of iterative layer pruning has achieved model compression from 8.03B parameters to 6.28B, 5.41B, and 4.54B parameters, after removing 8, 12, and 16 layers, respectively. Moreover, the quality degradation caused by pruning has been mitigated through fine-tuning on medium-sized data (80K-100K) and knowledge distillation. As demonstrated by Table \ref{tab:eval-transformers}, by the end of the process, the pruned model could recover most of the translation quality of the baseline model. For translation from Czech to German (CES-DEU), pruning 8 layers and then fine-tuning the resulting model retains 98\% of the translation quality (as measured by COMET) before knowledge distillation and 99\% after knowledge distillation. Interestingly, for translation from English to Egyptian Arabic (ENG-ARZ), the model resulting from pruning up to 16 layers and then fine-tuning outperforms the baseline model. This can be attributed to the initial quality of the baseline model for this language pair.

Moreover, we experimented with immediate recovery through fine-tuning the model after each pruning phase (i.e. pruning the fine-tuned 24-layer model into 20 layers instead of pruning the baseline model directly), and noticed that the final quality was similar to pruning the baseline directly and then only fine-tuning the pruned model. This matches the results demonstrated by \citet{Moslem2025-EfficientSpeech-IWSLT} who experimented with immediate fine-tuning after pruning each layer, and observed that this could lead to overfitting. In other words, it is sufficient to fine-tune the final pruned model.

In terms of inference performance, we observe that using \textit{vLLM}
\footnote{\scriptsize\url{https://github.com/vllm-project/vllm}} \cite{Kwon2023-vLLM} 
as an inference engine instead of \textit{Transformers} increases the inference speed by more than four times when conducting the evaluation on one A40 48GB GPU (cf.~Figure~\ref{fig:trasformers-vllm}).
Moreover, while 4-bit quantization using \textit{bitsandbytes} \cite{Dettmers2023-QLoRA} reduces the memory footprint, pruning results in higher inference speed and throughput (cf.~Table~\ref{tab:eval-arz-4bit}).

\section{Conclusions and Future Work}

In this work, we demonstrated that iterative layer pruning is an effective approach for compressing LLMs while retaining translation quality. The method relies on layer importance evaluation, followed by fine-tuning on a medium-sized dataset. This iterative layer pruning process reduces the model size and accelerates inference. To ensure reproducibility, we have made our code publicly available.\footnote{\scriptsize\url{https://github.com/ymoslem/Model-Compression}}

Future research directions include investigating adaptive compression approaches that dynamically select appropriate model configurations based on real-time deployment constraints such as memory limits and latency requirements.
Moreover, we plan to assess our compression methods on a broader range of datasets, including both sentence-level and document-level data. Since Aya-Expanse is designed to follow textual instructions, exploring retrieval-augmented generation combined with few-shot prompting presents a promising opportunity for enhancing translation performance in compressed models.

\section*{Acknowledgements}
We sincerely thank the ADAPT Centre (Ireland) and Kreasof AI (Indonesia) for providing the \mbox{resources} and support that made this work possible.

\bibliography{paperpile,other}

\appendix
\onecolumn

\section{Prompt for Synthetic Data Generation for Egyptian Arabic}
\label{app:egy-prompt}

\begin{small}
\begin{verbatim}
    I would like to convert a Standard Arabic text into Egyptian Arabic. 
    Please generate the Egyptian Arabic version using a neutral, informative 
    tone with slightly conversational phrasing, similar to the example below. 
    The output should feel natural, like it's written for a general Egyptian 
    audience but still accurate and clear. Do not add any commentary; just 
    return the Egyptian Arabic version.

    English:
    <english_example>

    Standard Arabic:
    <standard_arabic_example>

    Egyptian Arabic:
    <egyptian_arabic_example>

    English:
    {new_source_text}

    Standard Arabic:
    {new_target_text}

    Egyptian Arabic:
    
\end{verbatim}
\end{small}

\section{Evaluation of Egyptian Arabic Translation on WMT24++}

\begin{table}[htbp]
\centering
\begin{tabular}{@{}lccc@{}}
\toprule
\textbf{Model} & \textbf{Layers} & \textbf{chrF++} ↑ & \textbf{COMET} ↑ \\ 
\midrule
Baseline 32B & 40 & 33.89 & 75.55  \\
Baseline 8B  & 32 & 30.62 & 74.50 \\ 
\midrule
\multirow{3}{*}{Pruned + FT} & 24 & \textbf{37.01} & \textbf{76.86} \\ \cmidrule{2-4}
                             & 20 & 34.24 & 74.95  \\ \cmidrule{2-4}
                             & 16 & 29.32 & 68.70  \\
\bottomrule
\end{tabular}
\caption{Evaluation of the ENG-ARZ models fine-tuned with target-side synthetic data. The evaluation uses the WMT24++ benchmark and shows quality improvement compared to the baseline models.}
\label{tab:eval-arz-wmt}
\end{table}

\vspace{-10pt}
\section{Quantization Speed and Throughput}

\begin{table}[htbp]
\centering
\begin{tabular}{@{}lccccc@{}}
\toprule
\textbf{Model} & \textbf{Layers} & \textbf{4-bit} & \textbf{Memory} ↓ & \textbf{Speed} ↓ & \textbf{Throughput} ↑ \\ 
\midrule
\multirow{2}{*}{Baseline 8B} & \multirow{2}{*}{32} & no  & 14.96 & \uline{00:19} & \uline{2275} \\ 
                             &                     & yes & \hspace{5pt}\uline{5.61}  & 00:42 & 1053 \\ 
\midrule
\multirow{6}{*}{Pruned + FT} & \multirow{2}{*}{24} & no  & 11.71 & \uline{00:14} & \uline{3008} \\ 
                             &                     & yes & \hspace{5pt}\uline{4.70}  & 00:22 & 2004 \\ \cmidrule{2-6}
                             & \multirow{2}{*}{20} & no  & 10.08 & \uline{00:12} & \uline{3484} \\ 
                             &                     & yes & \hspace{5pt}\uline{4.24}  & 00:18 & 2367 \\ \cmidrule{2-6}
                             & \multirow{2}{*}{16} & no  & \hspace{5pt}8.46  & \uline{00:10} & \uline{4192} \\ 
                             &                     & yes & \hspace{5pt}\uline{3.78}  & 00:15 & 2908 \\
\bottomrule
\end{tabular}
\caption{Performance comparison of Aya-Expanse-8B baseline and the pruned models with and without \mbox{4-bit quantization}, in terms of memory (GiB), speed (mm:ss), and output throughput (tokens/sec). The evaluation uses the holdout ENG-ARZ News Commentary test dataset, on one A40 48GB GPU. While 4-bit quantization reduces the memory footprint, layer pruning achieves both higher inference speed and throughput.}\label{tab:eval-arz-4bit}
\end{table}


\end{document}